\documentclass[lettersize,journal]{IEEEtran}
\usepackage{amssymb,amsmath,amsfonts}
\usepackage{algorithmic}
\usepackage{algorithm}
\usepackage{multirow}
\usepackage{dsfont}
\usepackage{array}
\usepackage[caption=false,font=normalsize,labelfont=sf,textfont=sf]{subfig}
\usepackage{textcomp}
\usepackage{bbm}
\usepackage{color}
\usepackage{xcolor}
\usepackage{tabularx}
\usepackage{stfloats}
\usepackage{flushend}
\usepackage{url}
\usepackage{verbatim}
\usepackage{graphicx}
\usepackage{cite}
\usepackage[normalem]{ulem}
\usepackage{booktabs}
\usepackage[caption=false,font=normalsize,labelfont=sf,textfont=sf]{subfig}
\usepackage{wrapfig}
\newtheorem{proposition}{Proposition}

\newtheorem{theorem}{Theorem}

\newtheorem{definition}{Definition}
\newtheorem{assumption}{Assumption}
\newtheorem{setting}{Setting}

\begin{document}

\title{Cellular Traffic Prediction via Byzantine-robust Asynchronous Federated Learning}

\author{Hui~Ma, Kai~Yang, \IEEEmembership{Senior Member, IEEE}, and Yang~Jiao
\thanks{ This work was supported in part by the National Natural Science Foundation of China under Grant 12371519 and 61771013; in part by Asiainfo Technologies; in part by the Fundamental Research
Funds for the Central Universities of China; and in part by the Fundamental Research Funds of
Shanghai Jiading District. } \thanks{Hui Ma is with the School of Software, Xinjiang University, Kai Yang and Yang Jiao are with the Department of Computer Science, Tongji University, Shanghai 201804, China. (Corresponding author: Kai Yang: kaiyang@tongji.edu.cn)}}

\markboth{Journal of \LaTeX\ Class Files,~Vol.~14, No.~8, August~2021}
{Shell \MakeLowercase{\textit{et al.}}: A Sample Article Using IEEEtran.cls for IEEE Journals}

\maketitle

\begin{abstract}
Network traffic prediction plays a crucial role in intelligent network operation. Traditional prediction methods often rely on centralized training, necessitating the transfer of vast amounts of traffic data to a central server. This approach can lead to latency and privacy concerns. To address these issues, federated learning integrated with differential privacy has emerged as a solution to improve data privacy and model robustness in distributed settings. Nonetheless, existing federated learning protocols are vulnerable to Byzantine attacks, which may significantly compromise model robustness. Developing a robust and privacy-preserving prediction model in the presence of Byzantine clients remains a significant challenge. To this end, we propose an asynchronous differential federated learning framework based on distributionally robust optimization. The proposed framework utilizes multiple clients to train the prediction model collaboratively with local differential privacy. In addition, regularization techniques have been employed to further improve the Byzantine robustness of the models. We have conducted extensive experiments on three real-world datasets, and the results elucidate that our proposed distributed algorithm can achieve superior performance over existing methods.
\end{abstract}

\begin{IEEEkeywords}
Traffic prediction, federated learning, differential privacy, distributionally robust optimization.
\end{IEEEkeywords}

\section{Introduction}
\IEEEPARstart{W}{ith} the development of a range of emerging technologies \cite{Zhang2022}, beyond 5G and 6G technologies have garnered extensive interest from academia and industry \cite{Fadlullah2022,10043819} due to their ability to provide high data rates and reliability, and support a massive number of connected devices\cite{Qiu2023,jiao2020timeautoml}. Concurrently, the explosive growth of mobile devices and applications has significantly increased the size of wireless networks and the complicity of network traffic patterns\cite{10571806,8600752,8794857}. It will bring huge maintenance costs and workload to 6G technology, which drives the need for efficient nedtwork management and optimization. Network traffic prediction can alleviate network congestion and ensure stable network operation\cite{9293088}. Moreover, it enables efficient resources allocation, reduces maintenance costs, and improves the quality of service (QoS) and quality of experience (QoE)\cite{9930825} for users. Therefore, network traffic prediction techniques are significantly important\cite{ma2023cellular}.

 
Network traffic data exhibits complex temporal correlations and is susceptible to exogenous factors such as holidays and social events. However, existing network traffic prediction methods primarily account for auxiliary information such as holidays and weather while ignoring factors such as social activities and news events, resulting in poor performance. Besides, most existing traffic prediction methods rely on centralized deep learning methods that require transferring a large amount of raw data to a data center for learning a generalized prediction model. However, the raw data may contain sensitive information for individuals. If such sensitive information is leaked during the collection process, it will directly threaten the data security of individuals. To alleviate this problem, the federated learning framework based traffic prediction techniques have received much attention. In federated learning\cite{luo2023optimization}, the raw data of each local client does not need to be shared, which can protect data security to some extent.



However, in a federated learning framework, there remains a risk of privacy leakage even if only a subset of the model parameters or gradients are exchanged among local clients since they can infer data information from parameter updates \cite{Geiping2020}. Consequently, it is crucial to consider a privacy-preserving federated learning framework. Differential privacy\cite{Ha2019, Zhang2023} is one of the mainstream approaches that can provide privacy guarantees by injecting random noise into either the raw data or the model parameters of local clients, which ensures that statistical analysis results do not reveal information about individual data. Moreover, adding a large amount of random noise can enhance the privacy guarantee, but it will significantly reduce the accuracy of deep learning models. Therefore, it is a great challenge to design a federated learning framework based traffic prediction model that can strike a balance between data privacy and accuracy.

Recently, several studies modeled differential privacy based federated learning as a distributionally robust optimization problem\cite{Zhang2023}, owing to its ability to provide theoretical robustness guarantees. For instance, Shi et al.\cite{Shi2022} integrated local differential privacy with a federated learning framework to protect the privacy of local data. However, this method is computationally inefficient. It also uses a synchronized gradient update algorithm, which means its efficiency is constrained by the maximum communication and computational latency. In large-scale distributed systems, huge communication costs between clients can result in the unnecessary waste of resources (e.g., network bandwidth and computational power), which may reduce the efficiency and scalability of the entire system. Therefore, we employ an asynchronous gradient update algorithm to mitigate these issues. Furthermore, motivated by our previous research\cite{kaiyang2012,Yang2008}, this study adopt the distributiveness, which can be quantified by the amount of communication cost, to characterize the efficiency of our proposed algorithm.



On the other hand, in distributed machine learning, a minority of local clients may transmit incorrect gradients due to network latency or malicious attacks, potentially leading to significant disruptions in the learning process, known as Byzantine attacks. To alleviate this issue, several studies propose robust federated learning protocols such as robust aggregation rules (Krum\cite{Blanchard2017}, Median\cite{Yin2018}). However, these approaches often have high computational complexity and communication overhead. To reduce communication costs, \cite{MintongKang2024} utilized histogram statistics vectors to distinguish malicious clients while reducing privacy leakage. However, this study primarily focuses on Byzantine robustness and model performance, leaving a gap in the exploration of privacy level. The Byzantine robustness of federated deep learning models is the key to building reliable and trustworthy AI systems, and it remains a challenging task to design communication-efficient traffic prediction models with Byzantine robustness guarantees while ensuring the privacy of local client data and prediction accuracy. In recent years, some studies have utilized regularization techniques to develop communication-efficient federated learning models with Byzantine robustness \cite{Li2019,Zhu2022,Ma2022}. Inspired by this, our study employs regularization techniques to enhance the Byzantine robustness of our proposed algorithm while balancing privacy level and accuracy.

To address the above challenges, we propose a novel traffic prediction framework called Byzantine-robust asynchronous federated learning with differential privacy (BAFDP) algorithm. Specifically, \textit{to begin with}, we capture features from network traffic data and unstructured text data, which collects information of users' social activities and news events. Meanwhile, the previous research\cite{ma2024,wangtimemixer} indicates that network traffic data have short-term and periodic trends. Then, we model the long-term time-domain dependencies of traffic time series via capturing the short-term (hourly) and periodic (daily) features of network traffic data. \textit{Besides}, we design a federated learning framework based on local differential privacy, which involves adding random perturbations to the original local data and utilizing the noise data for distributed training. To characterize the uncertainty introduced by the random noise, we construct the uncertainty set using the Wasserstein distance, forming a Wasserstein ball centered at the empirical distribution $\tilde{\mathbb{D}}$ with a radius of the privacy level. \textit{Then}, we adopt an asynchronous gradient update algorithm that can reduce the computation time of local clients waiting for global model aggregation and reduce the communication delay. \textit{Finally}, we design an efficient and robust optimization algorithm to adjust the degree of Byzantine robustness for the prediction model using regularization techniques.

In detail, the contributions of this study are three-fold and can be summarized as follows:

\begin{itemize}
\item We design a robust privacy-preserving algorithm with superior prediction accuracy. To the best of our knowledge, this study is the first to explore the privacy-robustness-distributiveness-accuracy trade-off for network traffic forecasting.
\item We propose an asynchronous distributed algorithm to solve the robust optimization problem. It is a single-loop algorithm that is computationally efficient and fast convergence, which has not yet studied in cellular network traffic prediction scenarios.
\item Our study provides theoretical privacy guarantees and convergence analysis for the BAFDP algorithm, which can demonstrate the effectiveness of our proposed algorithm. Besides, we derive an upper bound for the convergence of our proposed algorithm, i.e., $T(\Upsilon) \sim O(\frac{1}{\Upsilon^2})$.
\end{itemize}

We conduct extensive experiments on three real-world datasets to validate the superior performance of BAFDP. The rest of this paper is organized as follows. We give a brief description of the related work in Section II. Then, we describe the preliminaries and problem definition in Section III and introduce the method in Section IV. After that, we present the detailed experimental setup and results in Section V and Section VI, respectively. Finally, we conclude our study in Section VII.

\section{Related work}
\subsection{Traffic Prediction based on Federated Learning}
In recent years, federated learning has gained widespread application in traffic prediction. For instance, Liu \textit{et al.} \cite{Liu2020a, Liu2020} proposed a federated learning algorithm based on gated recurrent units (FedGRU), while Sepasgozar \textit{et al.} \cite{Sepasgozar2022} introduced a federated learning framework for traffic flow prediction based on long and short-term memory networks (Fed-NTP). Moreover, Subramanya \textit{et al.} \cite{Subramanya2021} evaluated the prediction performance of centralized and federated learning frameworks for time series prediction models. In intelligent transportation, several studies \cite{Zhang2021a, Xia2022, Qi2022} designed a federated learning algorithm with graph neural networks for spatiotemporal traffic prediction. These studies adopt FedAvg algorithm to update local model parameters, which perform well when the local client data are independently and identically distributed (IID). However, the local client data is often non-independently identically distributed (non-IID), leading to poor performance of FedAvg on such data. To address the problem of data heterogeneity, Li \textit{et al.} \cite{Li2020b} introduced FedProx, which can be seen as a generalization and re-parameterization of FedAvg that includes a proximal term in the local objective function. Additionally, considering the varying contributions of multiple local clients to the global model, Ji \textit{et al.} \cite{Ji2018} proposed FedAtt, utilizing the attention mechanism to assign weights to each client. Zhang \textit{et al.} \cite{Zhang2021_fedDA} introduced a federated learning model based on a dual-attention mechanism (FedDA).

\subsection{Differential Privacy} 
Several studies \cite{Ouadrhiri2022, Ha2019} divided deep learning based differential privacy methods into three main types, including generating privacy-preserving datasets before training, adding noise to model parameters (or gradients) during training, and protecting user privacy during the model inference phase. During the training phase, Mireshghallah \textit{et al.} \cite{Mireshghallah2020} classified differential privacy into five categories based on the insert location of perturbations, including input-level, objective function-level, gradient update-level, output-level, and label-level perturbations. For instance, Huang \textit{et al.} \cite{9044400} added Laplacian noise to the edges and nodes of the graph structure. Besides, Shi \textit{et al.} \cite{Shi2022} generated a new privacy-preserving dataset to train a distributionally robust model based on federated learning, while Wang \textit{et al.} \cite{Wang2023} used synthetic data to train a global model, which can be considered a new privacy-preserving learning method. In addition, Shen \textit{et al.} \cite{Shen2022} incorporated regular terms into the objective function to improve the robustness of the training model. Regarding gradient update-level perturbations, \cite{Wei2020, Hu2020} added Gaussian noise to model parameters before sending them to the server for aggregation. Hao \textit{et al.} \cite{Hao2019} introduced a federated learning framework that combines differential privacy with homomorphic encryption. Zhang \textit{et al.} \cite{Zhang2023} proposed a robust game-theoretic federated learning framework with differential privacy, which can improve the defense against attacks using differential privacy by introducing incentives and compensation for the participants. Tavangaran \textit{et al.}  \cite{Tavangaran2022} proposed an iterative bi-level aggregation based federated learning framework to ensure privacy security for multiple base station scenarios. However, Ren \textit{et al.} \cite{10.1145/3510032} revealed that it is possible to recover the original local data (sensitive information) from the shared gradient in federated learning. 

\subsection{Byzantine Robustness} 
Many recent studies have explored the Byzantine robustness of deep learning models based on federated learning frameworks \cite{Guerraoui2023, Bouhata2022, ClippedClustering}. In general, it has been classified into three categories\cite{ClippedClustering}, including redundancy based methods, trust based schemes, and robust aggregation algorithms. For instance, \cite{Chen2018, Rajput2019} assigned redundant gradients to each local client and used them to eliminate the effects of Byzantine failures. However, the computational overhead is high when a large number of malicious clients exist. Regarding trust based schemes, Cao \textit{et al.} \cite{Cao2021} proposed FLTrust that requires a clean dataset (namely root dataset), which can be used to perform gradient updates on the server side. However, the root dataset is difficult to obtain in a large-scale distributed scenario. Although robust aggregation methods, such as Krum\cite{Blanchard2017}, GeoMed\cite{Chen2017}, AutoGM\cite{Li2023}, Centered Clipping\cite{Karimireddy2021} ClippedClustering\cite{ClippedClustering}, can improve Byzantine robustness without the need for root datasets, they have high computational complexity and training time in a large-scale distributed environment. Meanwhile, several studies \cite{Li2019, Ma2022} have proposed regularization based techniques to enhance the Byzantine robustness of deep learning models. For example, Li \textit{et al.} \cite{Li2019} proposed a robust stochastic aggregation (RSA) approach that combines a regularization term with an objective function. Furthermore, \cite{Zhu2022, Ma2022} proposed a robust model aggregation algorithm that can simultaneously address Byzantine robustness and ensure privacy preservation.

\subsection{Distributionally Robust Optimization}
By modeling uncertainty through uncertainty sets or probability distributions, distributionally robust optimization algorithms aim to find solutions that are resilient to uncertainty, ensuring optimal performance even under worst-case scenarios\cite{Jiao2022}. When robust optimization problems involve complex uncertainties, a common solution strategy is to transform them into bilevel optimization problems. Generally, there are three primary types of solutions. The first approach assumes that there exists an analytical solution to the inner-level optimization problem. Then, the bilevel optimization problem can be converted into a single-level optimization problem. However, it is difficult to find the analytical solution for the inner-level optimization problem.
In contrast, the second type of approach is gradient based techniques \cite{CANSAM2207}, which calculate the hyper-gradient (or hyper-gradient approximation estimate) and adopt gradient descent algorithms to solve the upper-level optimization problem \cite{Li2020c}. However, it is computationally intensive when computing hyper-gradient. The third type of method is the constraint based approach, which replaces the inner-level optimization problem with sufficient conditions for optimality. Then, the bilevel optimization problem can be transformed into a constrained single-level optimization problem \cite{Jiao2023}. For example, Mohri \textit{et al.} \cite{Mohri2019} introduced agnostic federated learning (AFL) and proposed a stochastic gradient descent algorithm for minimizing the maximum combination of empirical losses. Besides, Deng \textit{et al.} \cite{Deng2021} proposed a distributionally robust federated averaging algorithm, which uses a prior distribution to minimize the maximum loss in the worst-case distribution. However, it is a synchronous algorithm that is susceptible to the maximum delay clients, which will limit their application in large-scale distributed systems.

\section{Problem Formulation}
In this paper, we explore a distributed federated learning framework with a central server and $R$ local clients, which are composed of $M$ normal clients and $B$ Byzantine malicious clients, i.e., $R=M+B$. The Byzantine clients, as described in \cite{Zhu2022}, will collude with each other and send arbitrary malicious messages to the server. Additionally, the identity of malicious clients is a priori unknown to the server. Despite the presence of $B$ Byzantine malicious clients, our objective is to effectively utilize $M$ normal local clients for distributed training and obtain a Byzantine-robust global model. The diagram of our proposed algorithm (BAFDP) is shown in Figure 1.
\begin{equation}
\min _{\boldsymbol{\theta}} \sum \mathbb{E} \left[f_i\left(\boldsymbol{\theta}, \zeta_i\right)\right], i \in \{1,...,M\},
\end{equation}
where $\boldsymbol{\theta}$ represents the model parameters that need to be optimized. $f_i\left(\boldsymbol{\theta}, \zeta_i \right)$ denotes the loss function for the $i$-th local client with respect to the random variable $\zeta_i$.

\subsection{Differential Privacy}
In federated learning frameworks, a significant concern exists that deep learning models or gradients derived from local clients are at risk of privacy leakage, as an attacker might infer or reconstruct the training data from the learned models or gradients. Therefore, we implement differential privacy to protect the data privacy of local clients.

\textit{$(\epsilon,\delta)$-differential privacy.} Let $D_i$ denote the training samples and $D^{'}_i$ denote the dataset with only one training sample added or removed for the $i$-th client. The randomized mechanism $\mathcal{M}:\mathcal{X}\to \mathcal{R}$ with domain $\mathcal{X}$ and range $\mathcal{R}$ satisfies $(\epsilon,\delta)$-differential privacy, if for any two adjacent datasets $D_i,D^{'}_i\in \mathcal{X}$ and for any measurable sets $S \subseteq \mathcal{R}$, 
\begin{equation}
\mathbb{P}[\mathcal{M}(D_i) \in S] \leq e^{\epsilon} \mathbb{P}[\mathcal{M}(D^{'}_i) \in S]+\delta,
\end{equation}
where $\epsilon$ denotes the bound of all outputs on $D_i$ and $D^{'}_i$. $\delta$ is the ratio of the probabilities for $D_i$ and $D^{'}_i$ cannot be bounded by $e^{\epsilon}$ when we add a privacy preserving mechanism. A smaller $\epsilon$ provides a more ambiguous distinguish ability between neighboring datasets $D_i$ and $D^{'}_i$, thus increasing the level of privacy protection.

\subsection{Federated Learning based on Differential Privacy}
We assume that the privacy budget $\epsilon_{i}^t$ allocated to the $i$-th client at the $t$-th iteration does not exceed a predefined value $a$, where $T$ represents the total number of iterations. Consequently, we have the following constraint,
\begin{align}
& \ \ \  \ \ \ \epsilon_{i}^t \leq a, i \in \{1,... ,M\}, t\in \{1,...,T \}.
\end{align}

\begin{figure}[t]
\centering
\includegraphics[width=0.5 \textwidth]{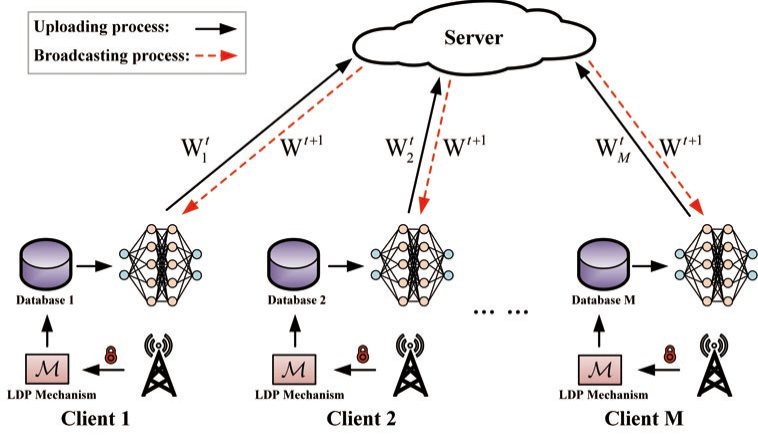} 
\caption{The diagram of BAFDP.}
\label{fig1}
\end{figure}

In our study, we consider the dataset for the $i$-th client $D_i=\{(\boldsymbol{x}_{i,j},\boldsymbol{y}_{i,j})\}_{j=1}^{N}$, where $\boldsymbol{x}_{i,j}=[\boldsymbol{x}_{i,j}^c,\boldsymbol{x}_{i,j}^p] \in \mathbb{R}^{d_x}$ represents the input sample and $\boldsymbol{y}_{i,j}\in \mathbb{R}^{d_y}$ denotes the corresponding target. Here, $\boldsymbol{x}_{i,j}^c$ and $\boldsymbol{x}_{i,j}^p$ refer to the short-term and period traffic data, $N$ is the number of samples, and $d_x,d_y$ are the dimensions of $\boldsymbol{x}_{i,j}$ and $\boldsymbol{y}_{i,j}$, respectively. Additionally, we define a probability distribution for the training samples for the $i$-th client as $\mathbb{D}_i$. In this paper, our primary focus is on multi-step ahead prediction, and we use $H$ to denote the number of steps predicted ahead.

Fig. 1 illustrates a federated learning framework incorporating a local differential privacy (LDP) mechanism for cellular traffic prediction. Specifically, Gaussian noise is introduced to each training sample. Consequently, the local data $D_i$ is transformed into $\tilde{D}_i=\{(\boldsymbol{\tilde{x}}_{i,j},\boldsymbol{y}_{i,j})\}_{j=1}^{N}$, where $\boldsymbol{\tilde{x}}_{i,j}=\boldsymbol{x}_{i,j}+\boldsymbol{v}_{i}^t$, and $\boldsymbol{v}_{i}^t \sim \mathcal{N}(0,\sigma^2_{i,t})$. Once the noise value is determined, each client perturbs its local data with Gaussian noise and subsequently trains the local model using the noisy data $\tilde{D}_i$. We denote the noisy data distribution of the $i$-th client as $\tilde{\mathbb{D}}_i$.

In federated learning, all participating clients conduct local training. By computing the empirical loss on the local training dataset $\tilde{D}_i$, we aim to obtain a global model with minimal loss expectation on the local data distribution,
\begin{align}
\mathbb{E}^{\tilde{\mathbb{D}}_i} \{\mathcal{L}(\boldsymbol{x},\boldsymbol{y},\boldsymbol{\omega}_i)\} = \frac{1}{N} \sum_{j=1}^{N}\mathcal{L}\left(g_{\boldsymbol{\omega}}(\boldsymbol{x}_{i,j}+\boldsymbol{v}_{i}^t,\boldsymbol{\omega}_i),\boldsymbol{y}_{i,j}\right),
\end{align}
where function $g_{\boldsymbol{\omega}}(*)$ denotes the neural network designed for cellular traffic prediction. It transforms the input $x_{i,j} + v_{i,t}$ into a predicted output through a series of nonlinear transformations, parameterized by the network weights $\boldsymbol{\omega}$. Additionally, $\boldsymbol{\omega}_i$ represents the learnable model parameters of the $i$-th client, and  $\mathcal{L}$ denotes the loss function used to evaluate the model's performance.

\subsection{Distributionally Robust Optimization} 
The model's utility is diminished due to the uncertainty introduced by local differential privacy noise. To alleviate this problem, we develop a distributionally robust optimization algorithm for traffic prediction, aiming to maximize the prediction performance of the model while adhering to the constraint of a limited privacy budget and providing robustness guarantees. More precisely, the distributionally robust optimization problem can be formulated as follows, 
\begin{align}
\min_{\boldsymbol{z} \in \mathcal{Z},\{\boldsymbol{w}_i \in \mathcal{W}\}, \{\epsilon_i^t \in \boldsymbol{\epsilon} \}} & \ \frac{1}{M} \sum_{i=1}^{M} \sup_{\tilde{\mathbb{D}}_{i}\in \mathcal{P}} \mathbb{E}^{\tilde{\mathbb{D}}_i} \{\mathcal{L}(\boldsymbol{x},\boldsymbol{y},\boldsymbol{\omega}_i)\},\label{Eq.6} \\
\text{s.t.} &  \ \  \epsilon_{i}^t \leq a,t \in \{1,...,T\},  \nonumber\\
&  \ \boldsymbol{z}=\boldsymbol{w}_i,i \in \{1,...,M\}, \nonumber \\
\text{var.}  &\  \boldsymbol{z},\boldsymbol{w}_1,\boldsymbol{w}_2,...,\boldsymbol{w}_M,\epsilon_{1}^t,...,\epsilon_{M}^t, \nonumber
\end{align}
where $\boldsymbol{z}$ is a global consensus variable that enables all local clients to converge on a consensus regarding the updated direction of the prediction model, thereby facilitating collaborative learning. $\mathcal{Z},\mathcal{W},\mathcal{P}$ are defined as non-empty closed convex set. The uncertainty set $\mathcal{P}$ consists of probability distributions generated from training samples.

\section{Method}
\subsection{Uncertainty Set}
To address the problem described in Eq. (\ref{Eq.6}), our initial step involves the construction of the uncertainty set $\mathcal{P}$. On the one hand, it should be rich enough to contain the traffic data distribution with a high degree of confidence. On the other hand, the uncertainty set has to be small enough to exclude pathological distribution that may lead to over-conservatism \cite{Esfahani2017}. In this study, we employ the Wasserstein distance as a metric to define the uncertainty set $\mathcal{P}$. More specifically, we construct a Wasserstein ball with radius $\rho_i^t$ around the empirical distribution $\tilde{\mathbb{D}}_i$. The optimal decision-making process is based on the worst-case distribution within the Wasserstein ball.

\begin{definition}\label{def.1}
($q$-Wasserstein distance) Consider a probability space $\Xi$ containing two samples $\boldsymbol{\xi}_1, \boldsymbol{\xi}_2$, each associated with the marginal probability distributions $\mathbb{P}_1$ and $\mathbb{P}_2$, respectively. For $q \geq 1$, $q$-Wasserstein\cite{Fournier2015} can be defined as,
\begin{equation}
\resizebox{0.9\hsize}{!}{$\begin{aligned}
\mathrm{W}_q (\mathbb{P}_1, \mathbb{P}_2) =  \inf \left\{ \left( \int_{\Xi\times \Xi} \| \xi_1- \xi_2 \|^q \pi (\mathrm{d} \xi_1, \mathrm{d} \xi_2) \right)^{\frac{1}{q}}   \right\} , 
\end{aligned} $}
\end{equation}
where $\pi \in \Pi \left(\mathbb{P}_1, \mathbb{P}_2\right)$, $\Pi$ denotes the joint probability distributions of two marginal probability distribution $\mathbb{P}_1,\mathbb{P}_2$, and $\|\cdot \|^q$ denotes the $q$-order norm function.
\end{definition}

As delineated in Definition \ref{def.1}, the computational complexity of $q$-Wasserstein distance escalates when value of $q$ becomes large. To circumvent this challenge, motivated by \cite{Farokhi2022}, our study opts to compute $1$-Wasserstein distance, wherein $q$ is equal to $1$. 

\begin{assumption}\label{ass.1}
(Light-tailed distribution \cite{Fournier2015}) There exists a $\beta > 1$ such that $E_{\xi \in \mathbb{D}}[e^{\|\xi \|^{\beta}}] < \infty$ if the distribution $\mathbb{D}$ is light-tailed.
\end{assumption}

Next, we aim to determine a suitable value for $\rho_i^t$ in order to construct an uncertainty set $\mathcal{P}$ that achieves a confidence level of $1-\gamma$. Then, it is necessary to assume that the data distribution $\mathbb{D}$ is a light-tailed distribution, as outlined in Assumption \ref{ass.1}. According to Lemma 2 from \cite{Shi2022}, we derive the radius $\rho_{i}^t$ using the following equation,
\begin{equation}
\rho_{i}^t = \eta_i + \sigma_{i,t},
\end{equation}
with 
\begin{equation}
\eta_i= \begin{cases}\left(\frac{\log \left(c_1 / \gamma\right)}{c_2 N}\right)^{1 / \max \{d, 2\}}, & N \geq \frac{\log \left(c_1 / \gamma\right)}{c_2}, \\ \left(\frac{\log \left(c_1 / \gamma\right)}{c_2 N}\right)^{1 / \beta}, & N<\frac{\log \left(c_1 / \gamma\right)}{c_2},\end{cases} 
\end{equation}
where $N$ represents the number of training samples, and $d=d_x+d_y$ with $d\neq 2$. $c_1,c_2$ are two positive values depend solely on the parameters $\beta$ and $d$. Furthermore, we employ Gaussian mechanism, i.e., $\sigma_{i,t} = \frac{c_3}{\epsilon_{i}^t}$, where $c_3 = \sqrt{2 d \text{log}(1.25/\delta)} \Delta$ (Theorem 1 from \cite{Farokhi2022}).

After determining the uncertainty set $\mathcal{P}$, the problem (\ref{Eq.6}) can be reformulated as,
\begin{align}
\min_{\boldsymbol{z} \in \mathcal{Z},\{\boldsymbol{w}_i \in \mathcal{W}\},\{\epsilon_i^t \in \boldsymbol{\epsilon} \}} & \frac{1}{M} \sum_{i=1}^{M} \ \sup_{\tilde{\mathbb{D}}_{i}^t:W_1(\mathbb{D}_{i},\tilde{\mathbb{D}}_{i}^t) \leq \rho_{i}^t} \mathbb{E}^{\mathbb{D}_i} \{\mathcal{L}(\boldsymbol{x},\boldsymbol{y},\boldsymbol{\omega}_i)\},\label{Eq.12}\\
\text{s.t.} &  \ \  \epsilon_{i}^t \leq a,t \in \{1,...,T\},  \nonumber\\
&  \ \boldsymbol{z}=\boldsymbol{w}_i,i \in \{1,...,M\}, \nonumber \\
\text{var.}  &\  \boldsymbol{z},\boldsymbol{w}_1,\boldsymbol{w}_2,...,\boldsymbol{w}_M,\epsilon_{1}^t,...,\epsilon_{M}^t, \nonumber
\end{align}

\begin{assumption}\label{ass.2}
(Lipschitz continuous\cite{Shi2022}) For any two samples $\xi_1,\xi_2$, the loss function $\mathcal{L}(\cdot)$ is $G(\boldsymbol{\omega})$-Lipschitz continuous, i.e., $|\mathcal{L}(\xi _1;\boldsymbol{\omega})-\mathcal{L}(\xi_2;\boldsymbol{\omega})| \leq G(\boldsymbol{\omega}) \cdot \|\xi_1-\xi_2 \|$. 
\end{assumption} 

The optimization problem as defined in Eq. (\ref{Eq.12}) involves finding an extreme value for the probability density function. It is computationally challenging to solve because we need to calculate all possible distributions. Inspired by \cite{Shi2022}, we transform the optimization problem (\ref{Eq.12}) into a more tractable reformulation. Prior to this transformation, we assume that the loss function $\mathcal{L}(\cdot)$ is Lipschitz continuous (Assumption \ref{ass.2}), a practical assumption in federated learning that facilitates theoretical analysis and ensures algorithmic stability\cite{Wei2020}.

\begin{proposition}\label{proposition.1}
If Assumption \ref{ass.2} holds, we can derive the following formulation,
\begin{equation}
\resizebox{0.95\hsize}{!}{$\begin{aligned}
\sup_{\tilde{\mathbb{D}}_{i}^t:W_1(\mathbb{D}_{i},\tilde{\mathbb{D}}_{i}^t) \leq \rho_{i}^t} \mathbb{E}^{\mathbb{D}_i} \{\mathcal{L}(\boldsymbol{x},\boldsymbol{y},\boldsymbol{\omega}_i)\} \leq \mathbb{E}^{\tilde{\mathbb{D}}_i^t} \{\mathcal{L}(\boldsymbol{x},\boldsymbol{y},\boldsymbol{\omega}_i)\}   + \rho_i^t \cdot G(\boldsymbol{\omega}_i), \label{Eq.13}
\end{aligned}$}
\end{equation}
\end{proposition}
\textit{Proof:}
\begin{align}
&\mathbb{E}^{\mathbb{D}_i} \{\mathcal{L}(\boldsymbol{x},\boldsymbol{y},\boldsymbol{\omega}_i)\} - \mathbb{E}^{\tilde{\mathbb{D}}_i}  \{\mathcal{L}(\boldsymbol{x},\boldsymbol{y},\boldsymbol{\omega}_i)\} \nonumber \\
&\leq \left| \mathbb{E}^{\mathbb{D}_i} \{\mathcal{L}(\boldsymbol{x},\boldsymbol{y},\boldsymbol{\omega}_i)\} 
- \mathbb{E}^{\tilde{\mathbb{D}}_i} \{\mathcal{L}(\boldsymbol{x},\boldsymbol{y},\boldsymbol{\omega}_i)\}  \right|. \label{Eq.14}
\end{align}

We assume that two samples $(\boldsymbol{x},\boldsymbol{y})$ and $(\boldsymbol{\tilde{x}},\boldsymbol{\tilde{y}})$ are obtained from the marginal probability distributions $\mathbb{D}_i$ and $\tilde{\mathbb{D}}_i$, respectively. Then, 
\begin{align}
 \ &\left| \mathbb{E}^{\mathbb{D}_i} \{\mathcal{L}(\boldsymbol{x},\boldsymbol{y},\boldsymbol{\omega}_i)\} - \mathbb{E}^{\tilde{\mathbb{D}}_i} \{\mathcal{L}(\boldsymbol{x},\boldsymbol{y},\boldsymbol{\omega}_i)\}  \right|, \nonumber
 \\
= &\left| \int_{\boldsymbol{x},\boldsymbol{y},\boldsymbol{\tilde{x}},\boldsymbol{\tilde{y}}} (\mathcal{L}(\boldsymbol{x},\boldsymbol{y},\boldsymbol{\omega}_i)-\mathcal{L}(\boldsymbol{\tilde{x}},\boldsymbol{\tilde{y}},\boldsymbol{\omega}_i)) \times \pi (\text{d}\boldsymbol{x},\text{d}\boldsymbol{y},\text{d}\boldsymbol{\tilde{x}},\text{d} \boldsymbol{\tilde{y}})   \right|, \nonumber \\
\leq &  \int_{\boldsymbol{x},\boldsymbol{y},\boldsymbol{\tilde{x}},\boldsymbol{\tilde{y}}} \left| \mathcal{L}(\boldsymbol{x},\boldsymbol{y},\boldsymbol{\omega}_i)-\mathcal{L}(\boldsymbol{\tilde{x}},\boldsymbol{\tilde{y}},\boldsymbol{\omega}_i) \right| \times \pi (\text{d}\boldsymbol{x},\text{d}\boldsymbol{y},\text{d} \boldsymbol{\tilde{x}},\text{d} \boldsymbol{\tilde{y}}) \nonumber\\
\leq & \int_{\boldsymbol{x},\boldsymbol{y},\boldsymbol{\tilde{x}},\boldsymbol{\tilde{y}}} \frac{\left| \mathcal{L}(\boldsymbol{x},\boldsymbol{y},\boldsymbol{\omega}_i)-\mathcal{L}(\boldsymbol{\tilde{x}},\boldsymbol{\tilde{y}},\boldsymbol{\omega}_i) \right|}{\|(\boldsymbol{x},\boldsymbol{y}) -(\boldsymbol{\tilde{x}},\boldsymbol{\tilde{y}}) \|} \times  \|(\boldsymbol{x},\boldsymbol{y}) -(\boldsymbol{\tilde{x}},\boldsymbol{\tilde{y}}) \| \nonumber \\
 &\ \ \ \ \ \ \ \ \ \ \  \ \ \ \ \ \ \ \ \ \  \ \ \ \ \ \ \ \ \ \ \ \ \ \ \ \ \ \ \ \ \ \ \ \times \pi (\text{d}\boldsymbol{x},\text{d}\boldsymbol{y},\text{d} \boldsymbol{\tilde{x}},\text{d} \boldsymbol{\tilde{y}}) \nonumber\\
\leq & G(\boldsymbol{\omega}_i) \cdot \rho_i. \label{Eq.15}
\end{align}

Then, we can obtain Eq. (\ref{Eq.13}) with the combination of Eq. (\ref{Eq.14}) and Eq. (\ref{Eq.15}). Furthermore, we convert the distributionally robust optimization problem (\ref{Eq.12}) to a regularized machine learning problem with the Lipschitz constant as a regularizer, i.e.,
\begin{align}
\min_{\boldsymbol{z} \in \mathcal{Z},\{\boldsymbol{w}_i \in \mathcal{W}\},\{\epsilon_i^t \in \boldsymbol{\epsilon} \} } & \frac{1}{M} \sum_{i=1}^{M} \left(\mathbb{E}^{\tilde{\mathbb{D}}_{i}^t} \{\mathcal{L}(\boldsymbol{x},\boldsymbol{y},\boldsymbol{\omega}_i)\} + \rho_{i}^t \cdot G(\boldsymbol{\omega}_i)\right). \label{Eq.16} \\
\text{s.t.} &  \ \  \epsilon_{i}^t \leq a,t \in \{1,...,T\},  \nonumber\\
&  \ \boldsymbol{z}=\boldsymbol{w}_i,i \in \{1,...,M\}, \nonumber \\
\text{var.}  &\  \boldsymbol{z},\boldsymbol{w}_1,\boldsymbol{w}_2,...,\boldsymbol{w}_M,\epsilon_{1}^t,...,\epsilon_{M}^t, \nonumber
\end{align}

We apply a first-order Taylor expansion to Eq. (\ref{Eq.16}), which aims to approximate the estimation of $\mathbb{E}^{\tilde{\mathbb{D}}_i^t} \{\mathcal{L}(\boldsymbol{x},\boldsymbol{y},\boldsymbol{\omega}_i)\}$, i.e.,
\begin{align}
&\mathbb{E}^{\tilde{\mathbb{D}}_{i}^t} \{\mathcal{L}(\boldsymbol{x},\boldsymbol{y},\boldsymbol{\omega}_i)\} =  \mathbb{E}^{\mathbb{D}_i} \{\mathcal{L}(\boldsymbol{x}_{i,j}+\boldsymbol{v}_{i}^t,\boldsymbol{y}_{i,j},\boldsymbol{\omega}_i)\} \nonumber\\
&\approx   \mathbb{E}^{\mathbb{D}_i} \{ \mathcal{L}(\boldsymbol{x}_{i,j},\boldsymbol{y}_{i,j},\boldsymbol{\omega}_i) + \boldsymbol{v}_{i}^t \nabla_{\mathcal{L}(\boldsymbol{x}_{i,j},\boldsymbol{y}_{i,j},\boldsymbol{\omega}_i)} \} \nonumber\\
&=   \mathbb{E}^{\mathbb{D}_i} \{ \mathcal{L}(\boldsymbol{x}_{i,j},\boldsymbol{y}_{i,j},\boldsymbol{\omega}_i)\} +\mathbb{E}^{\mathbb{D}_i} \{ \boldsymbol{v}_{i}^t\} \cdot \mathbb{E}^{\mathbb{D}_i}  \{\nabla_{\mathcal{L}(\boldsymbol{x}_{i,j},\boldsymbol{y}_{i,j},\boldsymbol{\omega}_i)}\} \nonumber\\
&=\frac{1}{N} \sum_{j=1}^{N} \mathcal{L}(\boldsymbol{x}_{i,j},\boldsymbol{y}_{i,j},\boldsymbol{\omega}_i), 
\end{align}
where $\boldsymbol{v}_{i}^t\sim \mathcal{N}(0,\sigma^2_{i,t})$, i.e., $\mathbb{E}^{\mathbb{D}_i} \{ \boldsymbol{v}_{i}^t\}=0$.

For brevity, we assume $g_{\boldsymbol{\omega}}(\boldsymbol{\omega}_i)=\frac{1}{N} \sum_{j=1}^{N} \mathcal{L}(\boldsymbol{x}_{i,j},\boldsymbol{y}_{i,j},\boldsymbol{ \omega}_{i})$. The robust optimization problem (\ref{Eq.16}) can be reformulated as,
\begin{align}
\min_{\boldsymbol{z} \in \mathcal{Z},\{\boldsymbol{w}_i \in \mathcal{W}\},\{\epsilon_{i}^t\in \boldsymbol{\epsilon} \} } & \frac{1}{M} \sum_{i=1}^{M}  \left( g_{\boldsymbol{\omega}}(\boldsymbol{\omega}_i) + (\eta_i + \frac{c_3}{\epsilon_{i}^t} ) \cdot G(\boldsymbol{\omega}_i) \right), \label{Eq.17}\\
\text{s.t.} &\ \ \epsilon_{i}^t \leq a,  \nonumber\\
& \ \ \ \boldsymbol{z}=\boldsymbol{w}_i,i \in \{1,...,M\}, \nonumber \\
\text{var.}  &\  \  \boldsymbol{z},\boldsymbol{w}_1,\boldsymbol{w}_2,...,\boldsymbol{w}_M,\epsilon_{1}^t,...,\epsilon_{M}^t. \nonumber
\end{align}

\subsection{BAFDP Algorithm}
Inspired by the approaches outlined in \cite{Li2019} and \cite{Zhu2022}, we introduce an $L1$-norm penalty term to the objective function, which can enable robust stochastic aggregation for our proposed model (Byzantine robustness). Consequently, the Augmented Lagrangian function corresponding to Eq. (\ref{Eq.17}) can be expressed as,
\begin{equation}
\begin{aligned}
\mathcal{L}_{\lambda} = &\frac{1}{M} \sum_{i=1}^{M} \Big( g_{\boldsymbol{\omega}}(\boldsymbol{\omega}_i) + (\eta_i + \frac{c_3}{\epsilon_{i}^t} ) \cdot G(\boldsymbol{\omega}_i) + \lambda_i (\epsilon_{i}^t-a) \\
&  +  \phi_i (\boldsymbol{z}-\boldsymbol{\omega}_i)+ \psi \| \boldsymbol{z}-\boldsymbol{\omega}_i \|_1 \Big), 
\end{aligned} \label{Eq.18}
\end{equation}
where $\mathcal{L}_{\lambda}=\mathcal{L}_{\lambda}(\{\boldsymbol{\omega}_i\},\{\epsilon_{i}^t\},\boldsymbol{z},\{\lambda_i\},\{\phi_i\})$, $\boldsymbol{\omega}_i$ and $\epsilon_i$ are decision variables, $\lambda_i$ and $\phi_i$ are the dual variables of inequality constraint and equality constraint in Eq. (\ref{Eq.17}), respectively. Additionally, $\psi$ represents a positive constant and $\psi \|\boldsymbol{z} -\boldsymbol{\omega}_i\|_1$ is the $L1$-norm penalty, whose minimization forces every $\boldsymbol{w}_i$ in close proximity to $\boldsymbol{z}$.

In our study, we utilize the regularized version of Eq. (\ref{Eq.18}), which can be written as,
\begin{equation}
\begin{aligned}
\bar{\mathcal{L}}_{\lambda} = \mathcal{L}_{\lambda} - \sum_{i=1}^{M} \Big(\frac{a_1^t}{2} \|\lambda_i \|^2 - \frac{a_2^t}{2} \|\phi_i \|^2\Big), 
\end{aligned}
\end{equation}
where $\bar{\mathcal{L}}_{\lambda}=\bar{\mathcal{L}}_{\lambda}(\{\boldsymbol{\omega}_i\},\{\epsilon_{i}^t\},\boldsymbol{z},\{\lambda_i\},\{\phi_i\})$, and $a_1^t,a_2^t$ represent the regularization terms.

In large distributed systems, synchronous distributed algorithms frequently encounter the "straggler" problem. The efficiency of the algorithm is limited by the worker with maximum communication and computation delay when the communication and computation capabilities of different local clients vary significantly. To address the issue, we propose an asynchronous federated learning algorithm, aiming at enhancing efficiency in solving robust optimization problems. To be specific, once the server receives model parameters from $S$ local clients, where $1 \leq S \leq M$. It updates model parameters $\boldsymbol{\omega}_i^{t}$, the privacy level $\epsilon_i^{t}$, the consensus variable $\boldsymbol{z}$, the dual variables $\lambda_i, \phi_i$. Specifically, during the $(t+1)$-th iteration, our proposed algorithm proceeds as follows.

\begin{algorithm}[tb] 
\caption{Byzantine-robust asynchronous federated learning with differential privacy (BAFDP) algorithm}
\label{alg:algorithm}
\textbf{Input}: Local datasets $D=\{D_1,D_2,...D_M\}$, the upper bound of privacy budget for each client $a$, the confidence level $1-\gamma$. \\
\textbf{Output}: Global model parameters $\boldsymbol{\omega}_i$, and the privacy level of differential privacy $\epsilon_i$.
\begin{algorithmic}[1] 
\STATE Randomly initialize variables $\{\boldsymbol{\omega}_i^0\}$, $\{\epsilon_{i}^0\}$, $\boldsymbol{z}^0$, $\{ \lambda^0_i \}$, and $\{\phi_i^0\}$. \;
\FOR{each iteration $t, t \in \{1,...,T\}$}
\FOR{each activate client $i,i\in \{1,...,S\}$}
\STATE Update parameters $\boldsymbol{\omega}_i^{t+1}$ and privacy constraint $\epsilon_{i}^{t+1}$ with Eq. (\ref{Eq.20}) and Eq. (\ref{Eq.21}), respectively;\\
\STATE Uploading $\boldsymbol{\omega}_i^{t+1}$ and $\epsilon_{i}^{t+1}$ to the server.
\ENDFOR
\STATE Server update parameters $\boldsymbol{z}^{t+1}$ and $\lambda^{t+1}_i$ with Eq. (\ref{Eq.22}) and Eq. (\ref{Eq.23}), respectively;
\STATE Server broadcasts $\boldsymbol{z}^{t+1}$ and $\lambda^{t+1}_i$ to activate clients. \\
\FOR{each activate client $i,i\in \{1,...,S\}$}
\STATE Update pairwise variable $\phi_i^{t+1}$ with Eq. (\ref{Eq.25}).
\ENDFOR
\ENDFOR
\STATE \textbf{return} $\boldsymbol{\omega}_i$, $\epsilon_{i}$
\end{algorithmic}
\end{algorithm}

\begin{definition}\label{def.3}
In the asynchronous distributed algorithm proposed in our study, during the $t$-th iteration, $\hat{t}$ is denoted as the last iteration when this client was activated and $\tilde{t}$ is designated as the subsequent iteration when the client is expected to be activated.
\end{definition}

\textit{Step 1.} Active clients update local model parameters $\boldsymbol{\omega}_i^t$ and $\epsilon_{i}^t$. Following Definition \ref{def.3}, we can obtain that $\boldsymbol{\omega}_i^{t}=\boldsymbol{\omega}_i^{\hat{t}}$, $\epsilon_{i}^{t}=\epsilon_{i}^{\hat{t}}$, and $\phi_i^{t}=\phi_i^{\hat{t}}$. Subsequently, $\boldsymbol{\omega}_i^{t+1}$ and $\epsilon_{i}^{t+1}$ can be updated using the following equations,
\begin{equation}
\begin{aligned}
\boldsymbol{\omega}_i^{t+1}&=\boldsymbol{\omega}_i^{t} - \alpha_{\boldsymbol{\omega}} \big(\nabla_{\boldsymbol{\omega}_{i}} \bar{\mathcal{L}}_\lambda(\{\boldsymbol{\omega}_i^{t}\},\{\epsilon_{i}^t\},\boldsymbol{z}^{t},\{\lambda^{t}_i\},\{\phi_i^t\} )  \\
& \ \ \ \ + \psi \text{sign}(\boldsymbol{z}_i^{t}-\boldsymbol{\omega}_i^{t}) \big),
\end{aligned}\label{Eq.20}
\end{equation}
\begin{equation}
\begin{aligned}
\epsilon_{i}^{t+1}&=\epsilon_{i}^t - \alpha_{\epsilon} \nabla_{\epsilon_{i}} \bar{\mathcal{L}}_\lambda(\{\boldsymbol{\omega}_i^{t+1}\},\{\epsilon_{i}^t\},\boldsymbol{z}^{t},\{\lambda^{t}_i\},\{\phi_i^{t}\}),
\end{aligned}\label{Eq.21}
\end{equation}
where $\alpha_{\boldsymbol{\omega}}$ and $\alpha_{\boldsymbol{\epsilon}}$ are the step sizes for updating variables. Once the local variables have been updated, the active clients proceed to transmit these updated variables to the central server.

\textit{Step 2.} Upon receiving updates from $S$ clients, the consensus variable $\boldsymbol{z}$ and the pairwise variable $\lambda_i$ will be updated in the central server with the following formula,
\begin{equation}
\resizebox{0.85\hsize}{!}{$\begin{aligned}
&\boldsymbol{z}^{t+1}=\boldsymbol{z}^t - \alpha_{\boldsymbol{z}} \Big( \nabla_{\boldsymbol{z}} \bar{\mathcal{L}}_\lambda(\{\boldsymbol{\omega}_i^{t+1}\},\{\epsilon_{i}^{t+1}\},\boldsymbol{z}^t,\{\lambda^t_i\},\{\phi_i^t\}) \label{Eq.22}  \\
& + \psi \big( \sum_{i \in \mathcal{R}} \text{sign}(\boldsymbol{z}^{t}-\boldsymbol{\omega}_i^{t+1}) + \sum_{j \in \mathcal{B}} \text{sign}(\boldsymbol{z}^{t}-\boldsymbol{\omega}_j^{t+1})\big)\Big), 
\end{aligned}  $}
\end{equation}
\begin{equation}
\resizebox{0.85\hsize}{!}{$\begin{aligned}
\lambda^{t+1}_i=\lambda^t_i + \alpha_{\lambda}\nabla_{\lambda_i} \bar{\mathcal{L}}_\lambda(\{\boldsymbol{\omega}_i^{t+1}\},\{\epsilon_{i}^{t+1}\},\boldsymbol{z}^{t+1},\{\lambda^t_i\},\{\phi_i^t\}),
\end{aligned}  $}\label{Eq.23}
\end{equation}
where $\alpha_{\boldsymbol{z}},\alpha_{\lambda}$ are step sizes. The sets $\mathcal{R}$ and $\mathcal{B}$ comprise $M$ normal clients and $B$ malicious clients, respectively. Subsequently, the central server broadcasts the updated variables to the active clients.

\textit{Step 3.} The local active clients update the pairwise variable $\phi_i$ with the following equation,
\begin{equation}
\resizebox{0.85\hsize}{!}{$\begin{aligned}
\phi_i^{t+1}=\phi_i^{t}+\alpha_{\phi}\nabla_{\phi_i} \bar{\mathcal{L}}_\lambda(\boldsymbol{\omega}_i^{t+1}\},\{\epsilon_{i}^{t+1}\},\boldsymbol{z}^{t+1},\{\lambda^{t+1}_i\},\{\phi_i^t\}), \label{Eq.25}
\end{aligned} $}
\end{equation}
where $\alpha_{\phi}$ represents the step size. A comprehensive description of this algorithm is presented in Algorithm 1.

\subsection{Convergence Analysis} 

In this section, we perform a convergence analysis to demonstrate the effectiveness of our proposed algorithm. To begin with, we introduce the following definitions and assumptions.

\begin{definition}\label{def.4}
($\Upsilon$-stationary point) When $\| \nabla F^t \|^2 \leq \Upsilon$ and $\Upsilon \geq 0$, $(\{\boldsymbol{\omega}_i^{t}\},\{\epsilon_{i}^t\},\boldsymbol{z}^{t},\{\lambda^{t}_i\},\{\phi_i^t\})$ is an $\Upsilon$-stationary point of the differentiable function $\mathcal{L}$, which can be written as $T(\Upsilon)=\text{min} \{t |\  \|\nabla F^t \|^2 \leq \Upsilon\}$.
\end{definition}

In addition, we take the following assumption on the loss functions in Eq. (\ref{Eq.18}), which have been widely adopted in\cite{ji2021bilevel,Ji2020ConvergenceOM}

\begin{assumption}\label{ass.3}
(Gradient Lipschitz\cite{ji2021bilevel}) We assume that the gradient of $\mathcal{L}_{\lambda}$ is Lipschitz continuous, denoted by the existence of $L>0$, such that,
\begin{equation}
\resizebox{1\hsize}{!}{$\begin{aligned}
& \left\|\nabla_\theta \mathcal{L}_{\lambda}\left(\{\boldsymbol{\omega}_i\},\{\epsilon_{i}\},\boldsymbol{z},\{\lambda_i\},\{\phi_i\}\right)-\nabla_\theta \mathcal{L}_{\lambda}\left(\{\boldsymbol{\bar{\omega}}_i\},\{\bar{\epsilon}_{i}\},\boldsymbol{\bar{z}},\{\bar{\lambda}_i\},\{\bar{\phi}_i\}\right) \right\| \\
& \leq L \left\|\left[\boldsymbol{\omega}_{\text {c}}-\boldsymbol{\bar{\omega}}_{\text{c}} ; \epsilon_{\text {c}}-\bar{\epsilon}_{\text{c}} ; \boldsymbol{z}-\boldsymbol{\bar{z}} ; \lambda_{\text {c}}-\bar{\lambda}_{\text {c}} ;  \phi_{\text {c}}-\bar{\phi}_{\text {c}}\right]\right\|, \label{e.7}
\end{aligned} $}
\end{equation}
where $\theta \in \{ \{\boldsymbol{\omega}_i\},\{\epsilon_{i}\},\boldsymbol{z},\{\lambda_i\},\{\phi_i\} \}$ and $[;]$ represents the concat operation. $\boldsymbol{\omega}_{\text {c}}-\boldsymbol{\bar{\omega}}_{\text{c}}=[\boldsymbol{\omega}_1-\boldsymbol{\bar{\omega}}_1; ...; \boldsymbol{\omega}_M-\boldsymbol{\bar{\omega}}_M]$, $\epsilon_{\text {c}}-\bar{\epsilon}_{\text{c}}=[\epsilon_{1}-\bar{\epsilon}_{1}; ...; \epsilon_{M}-\bar{\epsilon}_{M}]$, $\lambda_{\text {c}}-\bar{\lambda}_{\text {c}} = [\lambda_{1}-\bar{\lambda}_{1} ;...;\lambda_{M}-\bar{\lambda}_{M} ]$, $\phi_{\text {c}}-\bar{\phi}_{\text {c}} =[\phi_{1}-\bar{\phi}_{1} ;...;\phi_{M}-\bar{\phi}_{M} ]$.
\end{assumption}

\begin{assumption}\label{ass.4}
(Boundedness) Building on our previous studies\cite{Jiao2022,Jiao2023}, we assume that the variables are bounded for convenient analysis, i.e., $\|\boldsymbol{\omega}_i \|^2 \leq \mu_1$, $\|\epsilon_{i} \|^2 \leq \mu_2$, $\|\boldsymbol{z} \|^2 \leq \mu_1$, $\|\lambda_i \|^2 \leq \mu_3$,  $\|\phi_i^t \|^2 \leq \mu_4$. Besides, we suppose that before obtaining the $\Upsilon$-stationary point, the variables in the server satisfy that $\|\boldsymbol{z}^{t+1}-\boldsymbol{z}^{t}\|^2 + \|\lambda^{t+1}-\lambda^t \|^2  \geq \nu$, where $\nu > 0$ is a small constant. Moreover, for any $k \in \{ 1,..., \varkappa \}$, the change of the variables in the server is upper bounded within $\varkappa$ iterations, i.e., $\|\boldsymbol{z}^{t}-\boldsymbol{z}^{t-k}\|^2 \leq \varkappa k_0 \nu, \sum_{i=1}^{M}\|\lambda_i^{t}-\lambda_i^{t-k} \|^2 \leq \varkappa k_0 \nu$, where $k_0$ is a constant. 
\end{assumption}

\begin{setting}\label{setting.1}
We suppose $c_1^t,c_2^t$ are two nonnegative non-increasing sequences, i.e., $c_1^t=\frac{1}{{{\alpha_{\lambda}}^(t+1)}^{\frac{1}{4}}} \geq \underline{c}_1, c_2^t=\frac{1}{{{\alpha_{\phi}}^(t+1)}^{\frac{1}{4}}} \geq \underline{c}_2$, where $\alpha_{\lambda},\alpha_{\phi}$ are the size of steps. Also, $\underline{c}_1,\underline{c}_2$ satisfy that $0\leq \underline{c}_1 \leq \frac{1}{\alpha_{\lambda} c_0},0\leq \underline{c}_2 \leq \frac{1}{\alpha_{\lambda} c_0}$,$c=(\frac{16M^2}{\Upsilon^2}(\frac{\mu_{3}}{{\alpha_{\lambda}}^2}+\frac{\mu_{4}}{{\alpha_{\phi}}^2})^2 +1)^{\frac{1}{4}}$.
\end{setting}

\begin{theorem}\label{theorem.1}
(Iteration Complexity) We suppose step sizes $\alpha_{\boldsymbol{\omega}}=\alpha_{\epsilon}=\alpha_{\boldsymbol{z}}$=$\frac{2}{L+\alpha_\lambda M L^2+\alpha_{\phi} M L^2+8M \beta L^2\left(\frac{1}{\alpha_\lambda \underline{c}_1{ }^2}+\frac{1}{\alpha_{\phi} \underline{c}_2{ }^2}\right)}$, $\alpha_\lambda<\min \left\{\frac{2}{L+2 c_1^0}, \frac{1}{30 \varkappa k_0 N L^2}\right\} \text { and } \alpha_{\phi} \leq \frac{2}{L+2 c_2^0}$. Combined with Assumption \ref{ass.3} and Assumption \ref{ass.4} mentioned above, the iteration complexity of the proposed algorithm to obtain $\Upsilon$-stationary point is bounded by,
\begin{equation}
\begin{aligned}
&T(\Upsilon) \sim \mathcal{O}\Big(\max \Big\{\frac{16M^2}{\Upsilon^{2}} (\frac{ \mu_{3}}{{\alpha_{\lambda}}^{2}}+\frac{\mu_{4}}{\alpha_{\phi}{ }^{2}})^{2} , \nonumber\\ 
& (\frac{(4c_{6}+2\alpha_{\phi} L^{2} (M-S))\left(\bar{c}+k_{c} \varkappa(\varkappa-1)\right) c_{5}}{\Upsilon}+\sqrt{2})^{2}\Big\}\Big),
\end{aligned}\label{eq.95}
\end{equation}
where $\mu_{3},\mu_{4},\beta,\bar{c},c_6$ and $c_7$ are constants. 
\end{theorem}

We provide a detailed derivation process in the Appendix file\cite{auxiliary_file}, which consists of the following four steps. First, we derive Lemma 1 (see Appendix C) based on Assumption 3 and Assumption 4. Next, by combining Assumption 3, the Cauchy-Schwarz inequality, and Lemma 1, we obtain Lemma 2 (see Appendix D). Furthermore, leveraging the setting of $a_1,a_2,b_{1}, b_{3}$, along with Lemma 2, we derive Lemma 3 (see Appendix E). Finally, by integrating Definition 1 with the above three lemmas, we formally derive Theorem 1 (see Appendix F) for our proposed algorithm. 

It is seen from Theorem 1 that the upper bound of the iteration complexity of the proposed algorithm to achieve the $\Upsilon$-stationary point is $O(1/\Upsilon^2)$. The iteration complexity is influenced by the settings of some parameters in the proposed algorithm, e.g., $\Upsilon$, $S$,  $M$, where $S$ denotes the number of active workers in each iteration and $M$ represents the number of workers in federated learning framework. Specifically, the iteration complexity of our proposed algorithm will increase with a smaller $\Upsilon$ is required. Besides, the iteration complexity will decrease when we set a higher number of active workers in each iteration $S$ in the distributed algorithm. In addition, the iteration complexity of the proposed algorithm exhibits an exponential growth trend with the increase in the number of clients $M$ as well as the bounded values $\mu_{3}$ and $\mu_{4}$.

\section{Experimental Setup}
\subsection{Datasets} 
We leverage three real-world datasets (i.e., Milano, Trento, and LTE traffic data) to conduct comparative experiments on cellular traffic prediction methods.
\begin{itemize}
\item \textit{Milano}: It collects telco and textual data from Milan city, ranging from 1 November 2013 to 1 January 2014. Our study focuses on network traffic data at a time granularity of one hour, which can be obtained from Harvard Dataverse \cite{milan_data}. The textual data contains social pulse and daily news datasets, where the social pulse dataset \cite{Social_Pulse_milan} is made up of anonymized usernames, geographical coordinates, and geolocalized tweets. The daily news dataset \cite{milanotoday} contains titles, topics, and types of news information. 
\item  \textit{Trento}: The Trento dataset gathers telco data in Trentino city from 1 November 2013 to 1 January 2014. It is also publicly available \cite{trentino_data}. The textual data are also publicly available. The social pulse and website news data can be obtained through\cite{Social_Pulse_trentino,TrentoToday}, respectively.
\item \textit{LTE traffic}: The LTE traffic dataset is obtained from a private operator. It records the downlink traffic volume (in GB) from 28 December 2020 to 12 January 2021 with a temporal interval of 1 hour. 
\end{itemize}

\begin{table*}[t]
\renewcommand\arraystretch{1.25}
\centering
\caption{Prediction performance comparisons in terms of RMSE and MAE on three real-world datasets}
\label{tab:tab1}
\begin{tabular}{c|c|cc|cc|cc|c}
\hline
\multirow{2}{*}{Methods}     & \multirow{2}{*}{Metrics} & \multicolumn{2}{c|}{Milano}         & \multicolumn{2}{c|}{Trento}         & \multicolumn{2}{c|}{LTE traffic}  & \multirow{2}{*}{Average rank} \\ \cline{3-8}
                             &                          & H=1              & H=24             & H=1              & H=24             & H=1             & H=24            &                               \\ \hline
\multirow{2}{*}{FedGRU}      & RMSE                     & 59.7312          & 113.2298         & 63.9254          & 82.0859          & 0.6429          & 0.7103          & \multirow{2}{*}{8.08}         \\
                             & MAE                      & 35.5476          & 55.3787          & 27.4410          & 35.9906          & 0.4916          & 0.5351          &                               \\ \hline
\multirow{2}{*}{Fed-NTP}     & RMSE                     & 63.1627          & 116.0371         & 70.9524          & 83.1321          & 0.6377          & 0.7036          & \multirow{2}{*}{8.42}         \\
                             & MAE                      & 37.6146          & 54.8787          & 31.8112          & 36.3906          & 0.4901          & 0.5398          &                               \\ \hline
\multirow{2}{*}{FedAtt}      & RMSE                     & 67.5600          & 91.4277          & 47.8984          & 68.4910          & 0.5726          & 0.6522          & \multirow{2}{*}{5.58}         \\
                             & MAE                      & 33.8725          & 44.4130          & 21.3529          & 30.8320          & 0.4246          & 0.4802    &                               \\ \hline
\multirow{2}{*}{FedDA}       & RMSE                     & 68.8192          & 90.3906          & 44.6887          & 76.9570          & 0.5343          & 0.6711          & \multirow{2}{*}{5.83}         \\
                             & MAE                      & 34.2460          & 43.8103          & 21.7030          & 33.5177          & 0.3938          & 0.4992          &                               \\ \hline
\multirow{2}{*}{AFL}         & RMSE                     & 57.0544          & 103.5668         & 49.9386          & 60.9002          & 0.5396          & 0.6498          & \multirow{2}{*}{5.08}         \\
                             & MAE                      & 31.8624          & 54.3650          & 20.9850          & 27.4167          & 0.4003          & 0.4854          &                               \\ \hline
\multirow{2}{*}{ASPIRE-EASE} & RMSE                     & 59.3777          & 89.1233          & 42.1691          & 60.7578          & 0.5328          & 0.6496          & \multirow{2}{*}{3}            \\
                             & MAE                      & 31.0235          & \textbf{42.6280} & 18.9201          & 27.0827          & 0.3922          & 0.4803          &                               \\ \hline
\multirow{2}{*}{UDP}         & RMSE                     & 70.0903          & 88.3448          & 42.8287          & 57.6768          & 0.5687          & 0.6492          & \multirow{2}{*}{4.67}         \\
                             & MAE                      & 35.3238          & 44.1734          & 19.8049          & 28.7682          & 0.4267          & 0.4806          &                               \\ \hline
\multirow{2}{*}{NbAFL}       & RMSE                     & 58.8181          & 89.0849          & 39.6796          & 57.7409          & 0.5341          & 0.6581          & \multirow{2}{*}{3.25}         \\
                             & MAE                      & 30.4511          & 45.8976          & 17.5697          & 26.1434          & 0.3919          & 0.4813          &                               \\ \hline
\multirow{2}{*}{BAFDP}       & RMSE                     & \textbf{55.6839} & \textbf{87.9505} & \textbf{38.4066} & \textbf{54.5616} & \textbf{0.5289} & \textbf{0.6488} & \multirow{2}{*}{1.08}         \\
                             & MAE                      & \textbf{28.0826} & 43.5820          & \textbf{16.6767} & \textbf{25.6615} & \textbf{0.3893} & \textbf{0.4798} &                               \\ \hline
\end{tabular}
\end{table*}

\subsection{Baseline Methods} To evaluate the effectiveness of BAFDP, we conduct a comparative analysis of our proposed algorithm against eight existing approaches for prediction performance.

\begin{itemize}
\item \textbf {FedGRU}\cite{Liu2020a}: It is a federated learning based framework with GRU neural networks for traffic time series forecasting.
\item \textbf {Fed-NTP}\cite{Sepasgozar2022}: It is a federated learning algorithm for network traffic prediction using LSTM neural networks.
\item \textbf{FedAtt}\cite{Ji2018}: It is a federated learning framework with an attention mechanism to aggregate multiple local models.
\item \textbf{FedDA}\cite{Zhang2021_fedDA}: It is a wireless traffic prediction model with a dual attention based model aggregation scheme.
\item \textbf{AFL}\cite{Mohri2019}: It is an agnostic federated learning framework with agnostic loss to mitigate data heterogeneity.
\item  \textbf{ASPIRE-EASE}\cite{Jiao2022}: It is a distributionally robust federated algorithm that is communicationally efficient.
\item \textbf{UDP}\cite{wei2021user}: It develops a differential privacy algorithm with a federated learning framework that can balance convergence performance and privacy level.
\item  \textbf{NbAFL}\cite{Wei2020}: It is a federated learning framework based on differential privacy, where noise is added to weights before local models are aggregated into a global model.
\end{itemize}

\subsection{Evaluation Metrics}
We choose root mean square error (RMSE) and mean absolute error (MAE) as evaluation metrics to assess the performance of different methods for cellular traffic prediction\cite{ma2023cellular}. They are defined as, 
\begin{align}
\text{RMSE} &=  \sqrt{\frac{1}{N}\sum_{i=1}^{N}(\boldsymbol{y}_i-\boldsymbol{\hat{y}}_i)^{2}}, \\
\text{MAE} &= \frac{1}{N}\sum_{i=1}^{N}|\boldsymbol{y}_i-\boldsymbol{\hat{y}}_i|,
\end{align}
where $\boldsymbol{y}_i$ and $\boldsymbol{\hat{y}}_i$ denote the $i$-th target and predicted value, respectively.
 
\subsection{Experimental Details} 
\textbf{Preprocessing:} 
The Min-Max normalization is utilized to scale the numerical data such as cellular traffic, tweet coordinates, and the number of users, tweets, and articles to the interval of [0, 1]. Furthermore, we transform metadata such as holidays and the day of the week via one-hot encoding.

\textbf{Hyperparameters:} 
The grid search is performed to tune all adjustable hyperparameters. For other models, all baseline models use the parameters provided by the papers. As for our proposed model, we employ an Multilayer Perceptron (MLP) network for the experiments and utilize the Adam algorithm to train the model parameters.

We set the forecasting horizon as $H = \{1, 24\}$, corresponding to predicting traffic for the next hour and the next day, respectively. Additionally,  the test dataset comprises cellular traffic from the last seven days, while all traffic data from before is the training dataset. The optimal results are highlighted in boldface. All experiments are conducted on a Linux server with four 12 GB GPUs with NVIDIA TITAN X (Pascal).

\section{Experimental Results}
We conduct extensive experiments on three real-world datasets to evaluate the prediction performance of cellular traffic.

\subsection{Traffic Prediction Performance} 
Table \ref{tab:tab1} presents the prediction performance of nine approaches on three real-world datasets. Specifically, we rank the results for each approach and utilize the average rank to assess the prediction performance of different methods. It is observed that FedGRU and Fed-NTP perform poorly among the nine methods considered in our study. This is because they utilize FedAvg algorithm to update the gradients of the model parameters, which performs poorly on Non-IID data in large-scale distributed scenarios. Furthermore, FedAtt and FedDA assign weights to each local client individually, allowing them to address non-IID data issues. Consequently, they outperform FedGRU and Fed-NTP.

AFL is a robust optimization algorithm that learns a global model with consistently good performance on almost all clients by minimizing the distributionally robustness experience loss. However, it is difficult to adjust the robustness adaptively. Moreover, ASPIRE-EASE is an efficient, robust optimization algorithm that leverages a prior distribution to tackle the non-IID problem. Additionally, it allows flexible adjustment of the robustness degree through the constrained D-norm. As a result, it outperforms other baseline models and demonstrates optimal prediction performance. 

Furthermore, both UDP and NbAFL are federated learning frameworks incorporating differential privacy. In each iteration, a subset of local clients is selected for model training, and Gaussian noise is added to the mode parameters, which can be further aggregated in the server. Although they can theoretically protect user privacy, they have poorer prediction performance compared to ASPIRE-EASE, as adding noise will reduce the accuracy of the prediction model.

BAFDP is also a federated learning framework based on differential privacy. In contrast to UDP and NbAFL, BAFDP utilizes a robust optimization algorithm that enables the traffic prediction model to find the optimal decision, which can minimize the worst-case cost over the uncertainty set and achieve good prediction performance on the vast majority of clients. Consequently, the prediction performance of BAFDP is relatively optimal when compared to the state-of-the-art methods mentioned above.

\begin{figure}[t]
\centering
\includegraphics[width=\columnwidth]{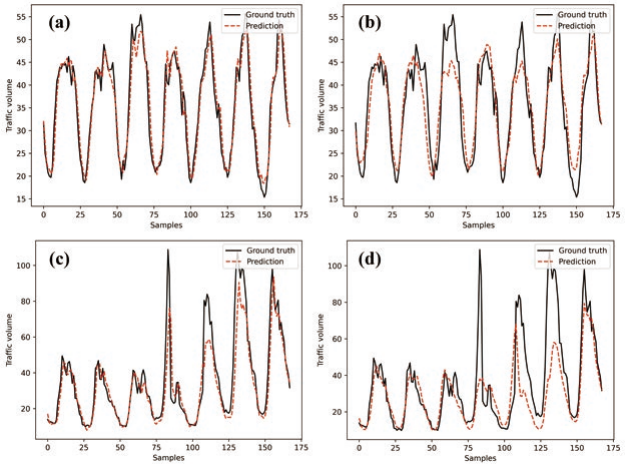} 
\caption{Visualization of prediction and ground truth on two open-sourced datasets.}
\label{fig2}
\end{figure}

In addition, we visualize the prediction results and ground truth values on two publicly available datasets to validate the algorithm’s prediction ability. The experimental results are shown in Fig. \ref{fig2}, where (a) and (b) are the results in one-hour-ahead prediction and one-day-ahead prediction evaluated on the Milano dataset, while (c) and (d) are the experimental results in one-hour-ahead prediction and one-day-ahead prediction evaluated on the Trento dataset, respectively. As can be observed from Fig. \ref{fig2} (a) that our proposed method can estimate the surge traffic accurately for the one-hour-ahead prediction on the Milano dataset. However, a very small fraction of the surge traffic could not be accurately predicted on the one-day-ahead prediction task. Meanwhile, we can obtain the similar  conclusion based on the experimental results on the Trento dataset.

\begin{table}[t]
\renewcommand\arraystretch{1.25}
\centering
\caption{Prediction performance in terms of privacy budget on the Milano dataset}
\label{tab:tab2}
\begin{tabular}{c|cc|cc}
\hline
\multirow{2}{*}{a} & \multicolumn{2}{c|}{H=1}            & \multicolumn{2}{c}{H=24}            \\ \cline{2-5} 
                   & RMSE             & MAE              & RMSE             & MAE              \\ \hline
10                 & 57.8359          & 29.0872          & 92.0481          & 43.9960          \\ \hline
20                 & 56.5829          & 28.5615          & 94.6091          & 45.6924          \\ \hline
30                 & 55.6839          & 28.0826          & 90.8518          & 46.1939          \\ \hline
40                 & 55.4800          & 27.8728          & \textbf{87.9505} & \textbf{43.5820} \\ \hline
50                 & \textbf{54.9889} & \textbf{27.7647} & 95.9304          & 45.5276          \\ \hline
60                 & 61.4517          & 30.3559          & 97.6415          & 46.0995          \\ \hline
70                 & 57.8373          & 29.0895          & 98.3936          & 46.5988          \\ \hline
\end{tabular}
\end{table}

\begin{table}[t]
\renewcommand\arraystretch{1.25}
\centering
\caption{Prediction performance in terms of privacy budget on the Trento dataset}
\label{tab:tab2.5}
\begin{tabular}{c|cc|cc}
\hline
\multirow{2}{*}{a} & \multicolumn{2}{c|}{H=1}            & \multicolumn{2}{c}{H=24}            \\ \cline{2-5} 
                   & RMSE             & MAE              & RMSE             & MAE              \\ \hline
0.1                & 38.6811          & 16.8104          & 59.5337          & 26.4041          \\ \hline
1                  & 38.4066          & 16.6767          & 54.5616          & 25.6615          \\ \hline
10                 & 38.3648          & 16.6632          & \textbf{46.1539} & \textbf{23.2244} \\ \hline
20                 & \textbf{38.3569} & \textbf{16.6541} & 57.8621          & 25.6610          \\ \hline
30                 & 40.2379          & 17.4964          & 58.5745          & 25.9706          \\ \hline
40                 & 39.4258          & 17.0954          & 58.5686          & 25.9498          \\ \hline
50                 & 39.0152          & 16.9005          & 58.9705          & 26.1497          \\ \hline
\end{tabular}
\end{table}

\begin{table}[t]
\renewcommand\arraystretch{1.25}
\centering
\caption{Prediction performance comparisons on the Milano dataset}
\label{tab:tab3}
\begin{tabular}{c|c|cc|cc}
\hline
\multirow{2}{*}{Models} & \multirow{2}{*}{Ratio} & \multicolumn{2}{c|}{H=1}            & \multicolumn{2}{c}{H=24}            \\ \cline{3-6} 
                        &                               & RMSE             & MAE              & RMSE             & MAE              \\ \hline
RSA                     & 0.1                           & 56.9244          & 29.7902          & 97.5367          & 51.1615          \\ \hline
DP-RSA                  & 0.1                           & 57.5793          & 30.6590          & 97.4712          & 51.6784          \\ \hline
\multirow{3}{*}{BAFDP}  & 0                             & 55.6839          & \textbf{28.0826} & \textbf{87.9505} & \textbf{43.5820} \\
                        & 0.1                           & \textbf{55.5178} & 27.9756          & 96.0341          & 45.9593          \\
                        & 0.3                           & 59.1019          & 29.8287          & 105.2124         & 48.7995          \\ \hline
\end{tabular}
\end{table}

\begin{figure}[t]
\centering
\includegraphics[width=\columnwidth]{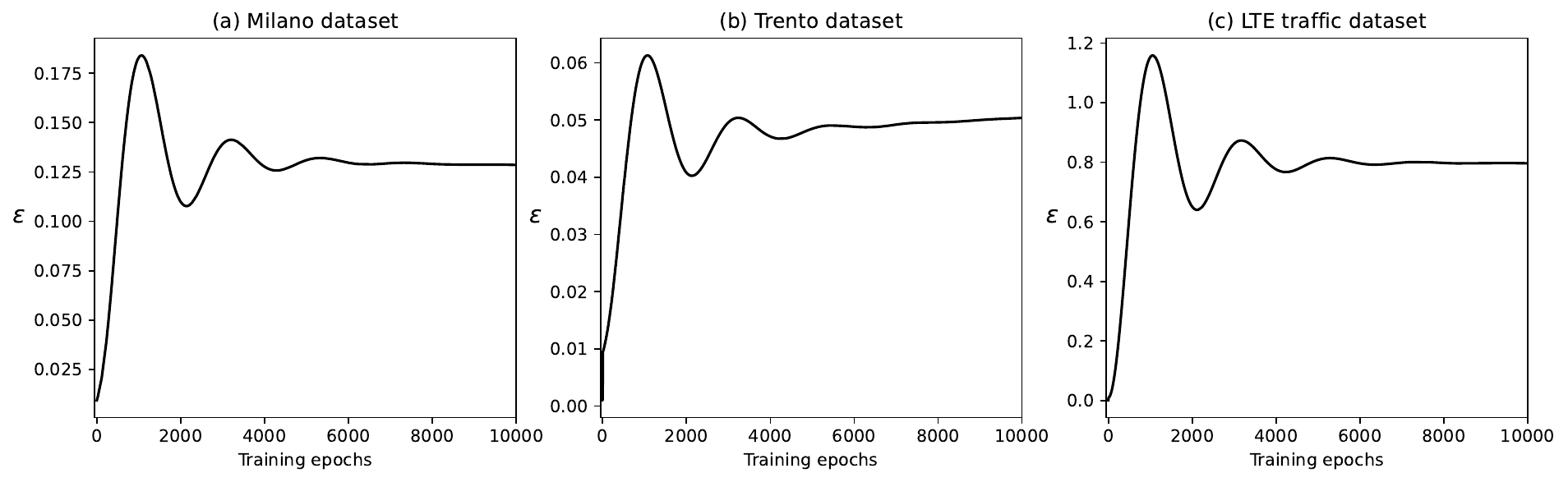} 
\caption{Visualization of privacy level changes during the training process on three real-world datasets.}
\label{fig3}
\end{figure}

\subsection{Privacy Level Analysis} 
For a better illustration, we visualize the trend of the privacy level $\boldsymbol{\epsilon}$ with the increasing number of iterations on three real-world datasets, including (a) the Milano dataset, (b) the Trento dataset, and (c) the LTE traffic dataset. Specifically, we present the experimental results of a randomly selected local client for each dataset in a one-hour-ahead prediction. It can be observed from Fig. \ref{fig3} that the privacy level $\boldsymbol{\epsilon}$ exhibits a significant upward trend as the number of iterations gradually increases to 1000. Subsequently, it demonstrates an oscillating to stable trend with further increases in the number of iterations. For instance, the privacy levels of a local client on the Milano dataset converge to 0.126. Our proposed algorithm can also provide different privacy levels for distinct local clients. The visualization facilitates observing how the privacy level evolves for each local client changes throughout the training process.

\subsection{Privacy Budget Analysis} 
We investigate the impact of the privacy budget $a$ on the one-hour-ahead and one-day-ahead prediction performance in terms of the Milano and Trento datasets. The results are shown in Table \ref{tab:tab2} and Table \ref{tab:tab2.5}. It can be seen from Table \ref{tab:tab2} that when performing the one-hour-ahead prediction task on the Milano dataset, the prediction accuracy of the global model gradually improved when the privacy budget increased from 10 to 50. However, when the privacy budget exceeds 50, it is difficult to get further improvement in the model prediction accuracy. Similarly, we can achieve the optimal prediction performance on the one-hour-ahead forecasting when the privacy budget reaches 20 on the Trento dataset, which is shown in Table \ref{tab:tab2.5}. Therefore, the appropriate privacy budget needs to be carefully chosen to obtain a better global model.

\subsection{Training Efficiency Analysis} 

\begin{figure}[t]
\centering
\includegraphics[width=\columnwidth]{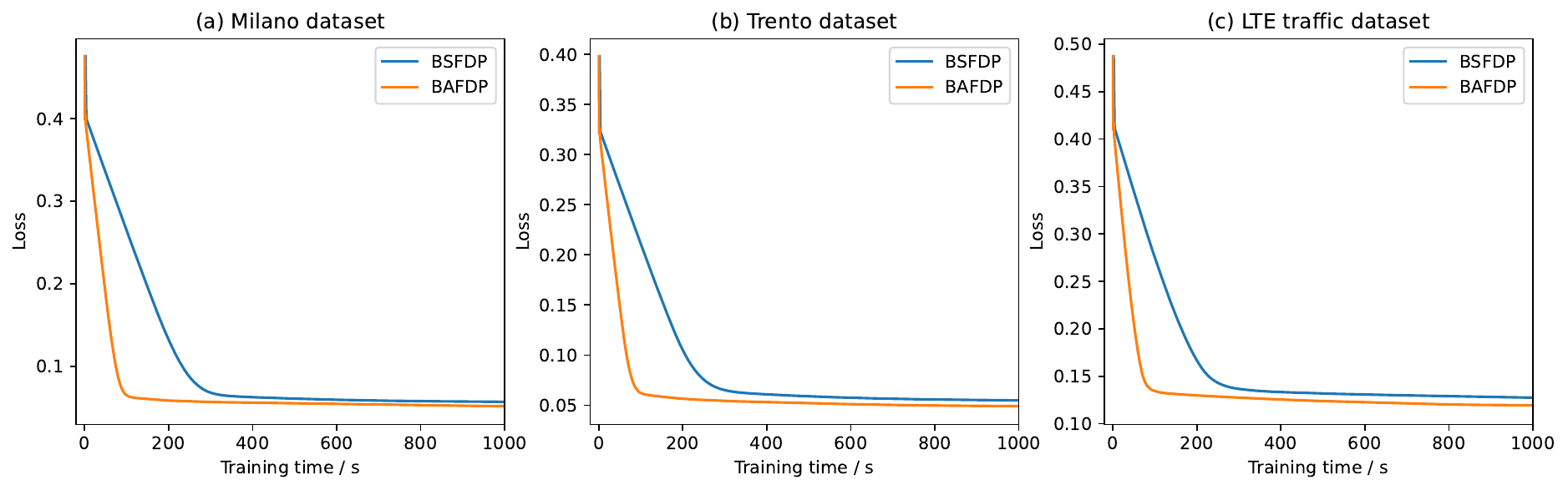} 
\caption{Comparison of training loss for synchronous and asynchronous distributed algorithm on three real-world datasets.}
\label{fig3}
\end{figure}

\begin{figure}[t]
\centering
\includegraphics[width=\columnwidth]{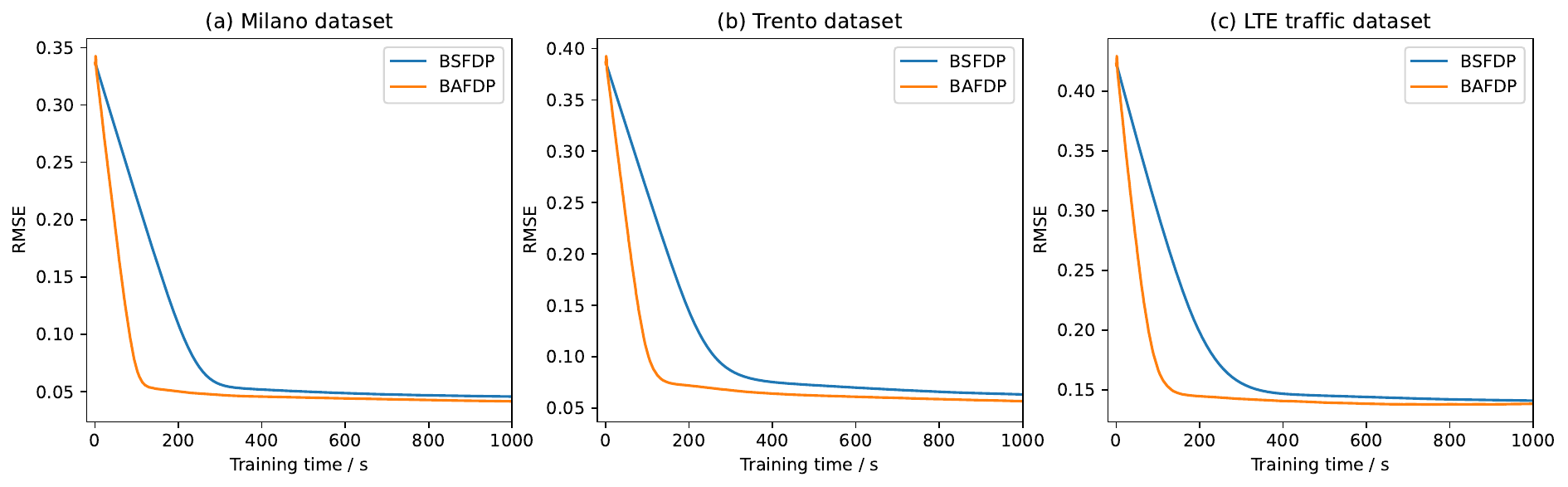} 
\caption{Comparison of RMSE for synchronous and asynchronous distributed algorithm on three real-world datasets.}
\label{fig4}
\end{figure}

\begin{figure}[t]
\centering
\includegraphics[width=\columnwidth]{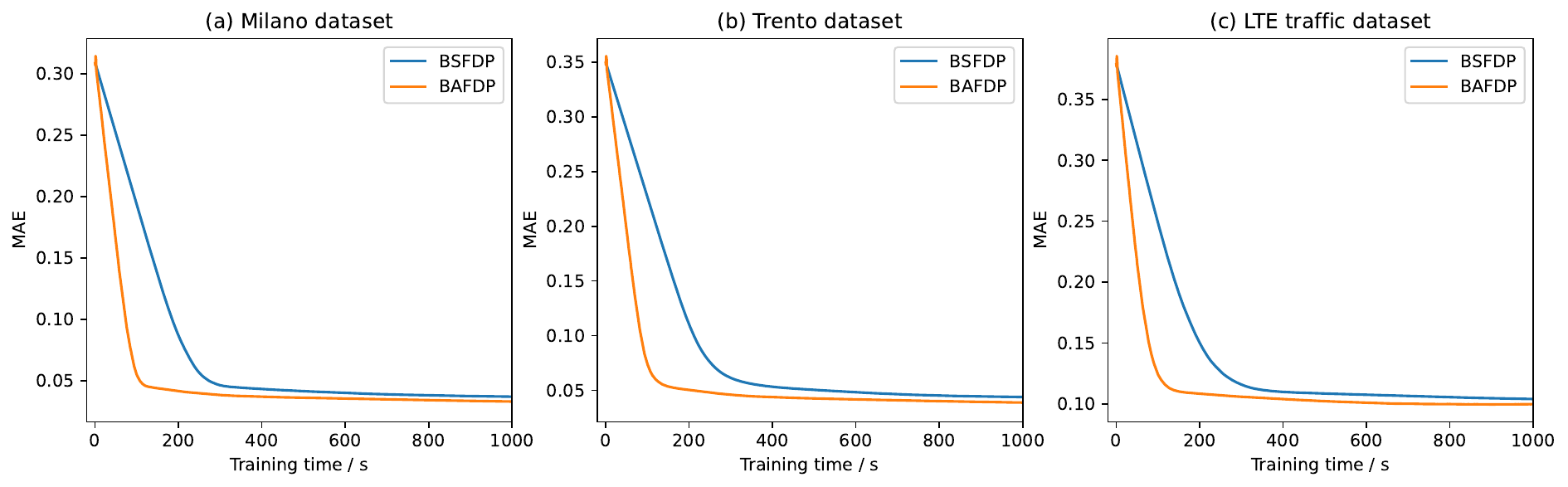} 
\caption{The comparison of MAE for synchronous and asynchronous distributed algorithm on three real-world datasets.}
\label{fig5}
\end{figure}

We conduct empirical comparisons of synchronous and Asynchronous distributed algorithms on three real-world datasets, including (a) the Milano dataset, (b) the Trento dataset, and (c) the LTE traffic dataset. Specifically, experiments are performed for the one-hour-ahead prediction. We denote the synchronous and asynchronous robust federated learning with differential privacy as BSFDP and BAFDP, respectively. Fig. \ref{fig3} illustrates the changes in training loss for SRFDP and BAFDP over increasing training time, while Fig. \ref{fig4} and Fig. \ref{fig5} depict how the evaluation metrics (RMSE and MAE) of BSFDP and BAFDP on the test dataset change with increasing training time, respectively.

It can be observed from Fig. (\ref{fig3}), Fig. (\ref{fig4}), and Fig. (\ref{fig5}) that our proposed BAFDP requires less training time to converge. This is because the server must wait for all local clients to update the model parameters in SRFDP. When some local clients lag behind, the synchronized overhead becomes significant, potentially reducing overall training performance. In contrast, BAFDP processes gradients from each local client asynchronously. Specifically, the server receives gradients from each local client, updates the global model, and immediately sends the updated global model back to the server. In this way, the asynchronous approach allows each local client to proceed to the next iteration without waiting for others. Consequently, our study employs an asynchronous distributed algorithm for gradient updates, which can reduce the computation time of the global model waiting for local clients, decrease communication delays, and enhance computational efficiency.

\subsection{Byzantine Robustness Analysis}
In this section, we investigate the impact of different Byzantine-robust levels on prediction accuracy, distributiveness and convergence rate of our proposed algorithm. For better illustration, we conduct experiments on the one-step-ahead prediction in terms of the Milano dataset.

\subsubsection{Accuracy}
We explore the impact of diverse Byzantine-robust levels on prediction accuracy by employing Byzantine-robust federated learning methods based on regularization techniques, i.e., RSA\cite{Li2019} and DP-RSA\cite{Zhu2022}. The experimental results are presented in Table \ref{tab:tab3}. As seen in this table, when the ratio of malicious clients is set to 0.1, RSA outperforms DP-RSA in terms of prediction accuracy. This is because DP-RSA introduces random perturbations to model gradients, consequently decreasing prediction accuracy. Additionally, our proposed algorithm demonstrates superior prediction accuracy compared to DP-RSA. This is because the privacy level in DP-RSA needs to be set manually. In contrast, our proposed algorithm optimizes model parameters and privacy levels simultaneously, enhancing prediction accuracy. Furthermore, we can observe a decreasing trend in prediction accuracy for our proposed algorithm as the percentage of malicious clients increases.

\subsubsection{Distributiveness}
\begin{figure}[t]
\centering
\includegraphics[width=0.7\columnwidth]{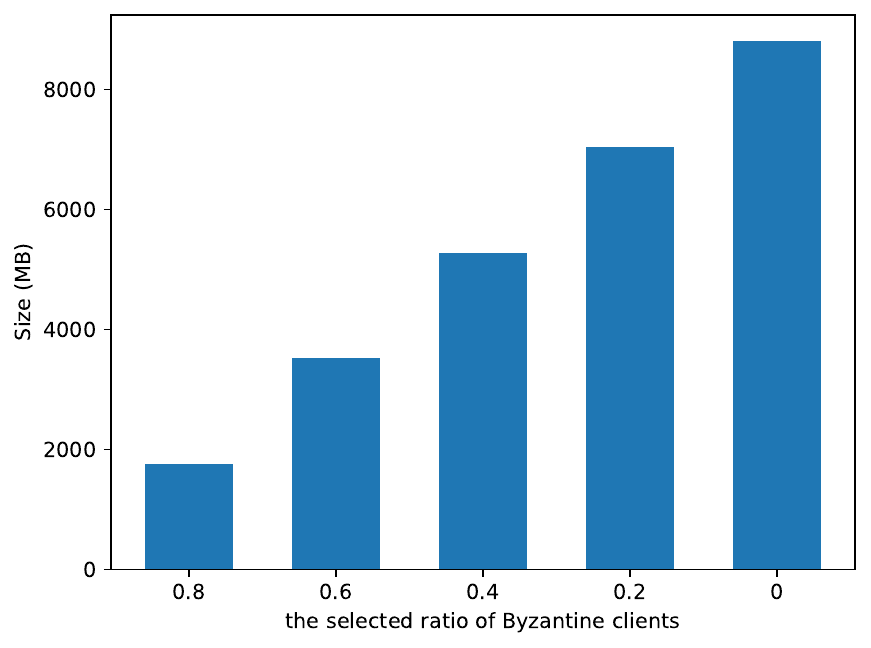} 
\caption{Size to be transmitted in federated learning framework in terms of different robustness levels.}
\label{fig3-3}
\end{figure}
In federated learning, the distributiveness of algorithms is related to the number of clients involved in training, the model size, and the number of iteration rounds. For better illustration, we consider the multilayer perceptron (MLP) model (with the size of 440MB) with 10,000 iterations, and set a fraction of malicious clients, i.e., $\{0.2,0.4,0.6,0.8,1\}$ from 10 local clients for experiments. In each iteration round, the size to be transferred is 2 $\times $ model size $\times $ participants as the model weights are broadcast both from the server to local clients and vice versa. Fig. \ref{fig3-3} illustrates the relationship between Byzantine robustness and distributiveness. As the proportion of malicious clients decreases, the number of local clients engaged in training gradually increases, leading to a proportional linear growth in the distributiveness of our proposed algorithm.

\subsubsection{Coverage rate}
\begin{figure}[t]
\centering
\includegraphics[width=0.7\columnwidth]{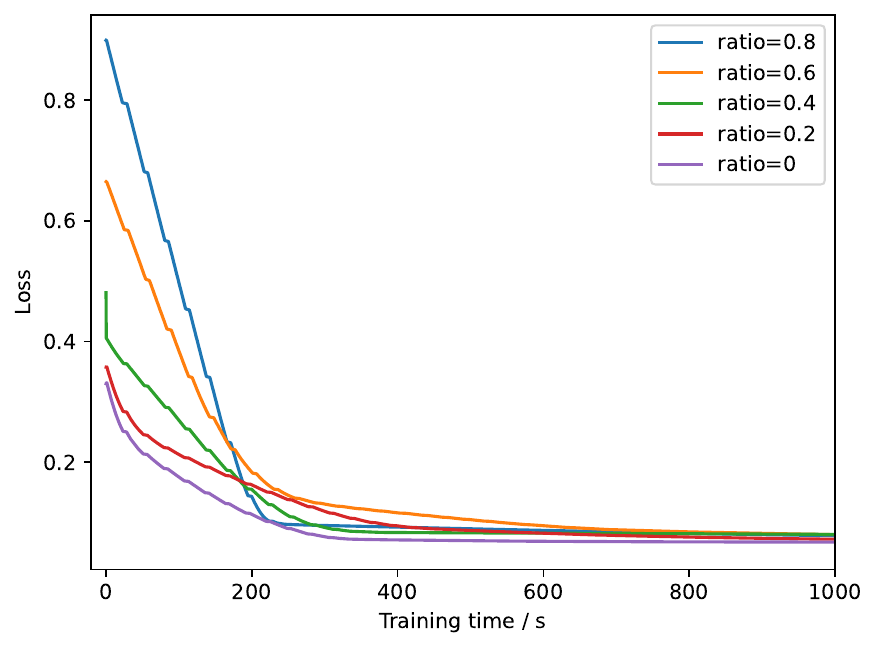} 
\caption{the training loss within different robustness levels}
\label{fig3-2}
\end{figure}
This section investigates the impact of different Byzantine-robust levels on the training loss. Specifically, we set the proportion of malicious clients as 0.8, 0.6, 0.4, 0.2, and 0, respectively. The experimental results are shown in Fig \ref{fig3-2}, from which it can be observed that the training time for convergence diminishes as the ratio of malicious clients decreases. It is widely acknowledged that a smaller ratio of malicious clients corresponds to a higher proportion of normal clients when the total number of local clients remains constant. Then, as the number of local clients involved in training keeps increasing, the training time required to achieve convergence is proportionally reduced. Consequently, the convergence rate is accelerated.

\section{Conclusion}
This work presents a Byzantine-robust asynchronous federated learning framework with differential privacy for cellular traffic prediction. We construct the perturbed training data with the local differential privacy mechanism and formulate the cellular traffic prediction as a robust optimization problem. Then, we design an asynchronous distributed algorithm to optimize model parameters and privacy levels simultaneously. Furthermore, we conduct extensive experiments on three real-world datasets. The results show the effectiveness of our proposed method, which outperforms other state-of-the-art approaches for cellular traffic forecasting.

In future work, we plan to explore the application of the federated learning framework based on differential privacy in diverse domains such as intelligent transportation and electricity. Concurrently, investigating the design of generalized time-series large language models for intelligent network operation and real-time prediction is another research direction to be carried out in the future.

\bibliographystyle{IEEEtran}
\bibliography{refs}

\begin{IEEEbiography}
	[{\includegraphics[width=1in,height=1.25in,clip,keepaspectratio]{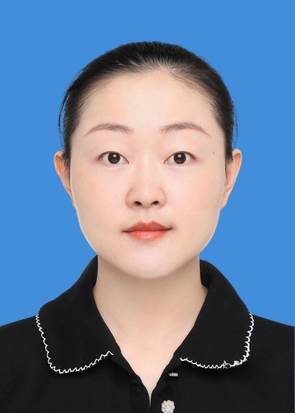}}]{Hui Ma} was born in Xinjiang, China. She
	received B.Eng. degree and M.S. degree from Jiangnan University, Wuxi, China, and Ph.D. degree from Tongji University, Shanghai, China. 
	
	She is a lecturer with Xinjiang University, Xinjiang, China. Her current research interests primarily revolve around time series data analytics, deep learning, spatiotemporal traffic forecasting, federated learning algorithms.
\end{IEEEbiography}

\begin{IEEEbiography}
	[{\includegraphics[width=1in,height=1.25in,clip,keepaspectratio]{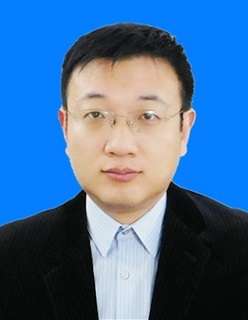}}]{Kai Yang} (Senior Member, IEEE) 
	received the B.Eng. degree from Southeast University, Nanjing, China, the M.S. degree from the National University of Singapore, Singapore, and the Ph.D. degree from Columbia University, New York, NY, USA.
	
	He is a Distinguished Professor with Tongji University, Shanghai, China. He was a Technical Staff Member with Bell Laboratories, Murray Hill, NJ, USA. He has also been an Adjunct Faculty Member with Columbia University since 2011. He holds over 20 patents and has been published extensively in leading IEEE journals and conferences. His current research interests include big data analytics, machine learning, wireless communications, and signal processing.
	
	Dr. Yang was a recipient of the Eliahu Jury Award from Columbia University, the Bell Laboratories Teamwork Award, the Huawei Technology Breakthrough Award, and the Huawei Future Star Award. The products he has developed have been deployed by Tier-1 operators and served billions of users worldwide. He is an Editor for the IEEE INTERNET OF THINGS JOURNAL, IEEE COMMUNICATIONS SURVEYS \& TUTORIALS, and a Guest Editor for the IEEE JOURNAL ON SELECTED AREAS IN COMMUNICATIONS. From 2012 to 2014, he was the Vice-Chair of the IEEE ComSoc Multimedia Communications Technical Committee. In 2017, he founded and served as the Chair of the IEEE TCCN Special Interest Group on AI Embedded Cognitive Networks. He has served as a Demo/Poster Co-Chair of IEEE INFOCOM, Symposium Co-Chair of IEEE GLOBECOM, and Workshop Co-Chair of IEEE ICME.
\end{IEEEbiography}

\begin{IEEEbiography}[{\includegraphics[width=1in,height=1.25in,clip,keepaspectratio]{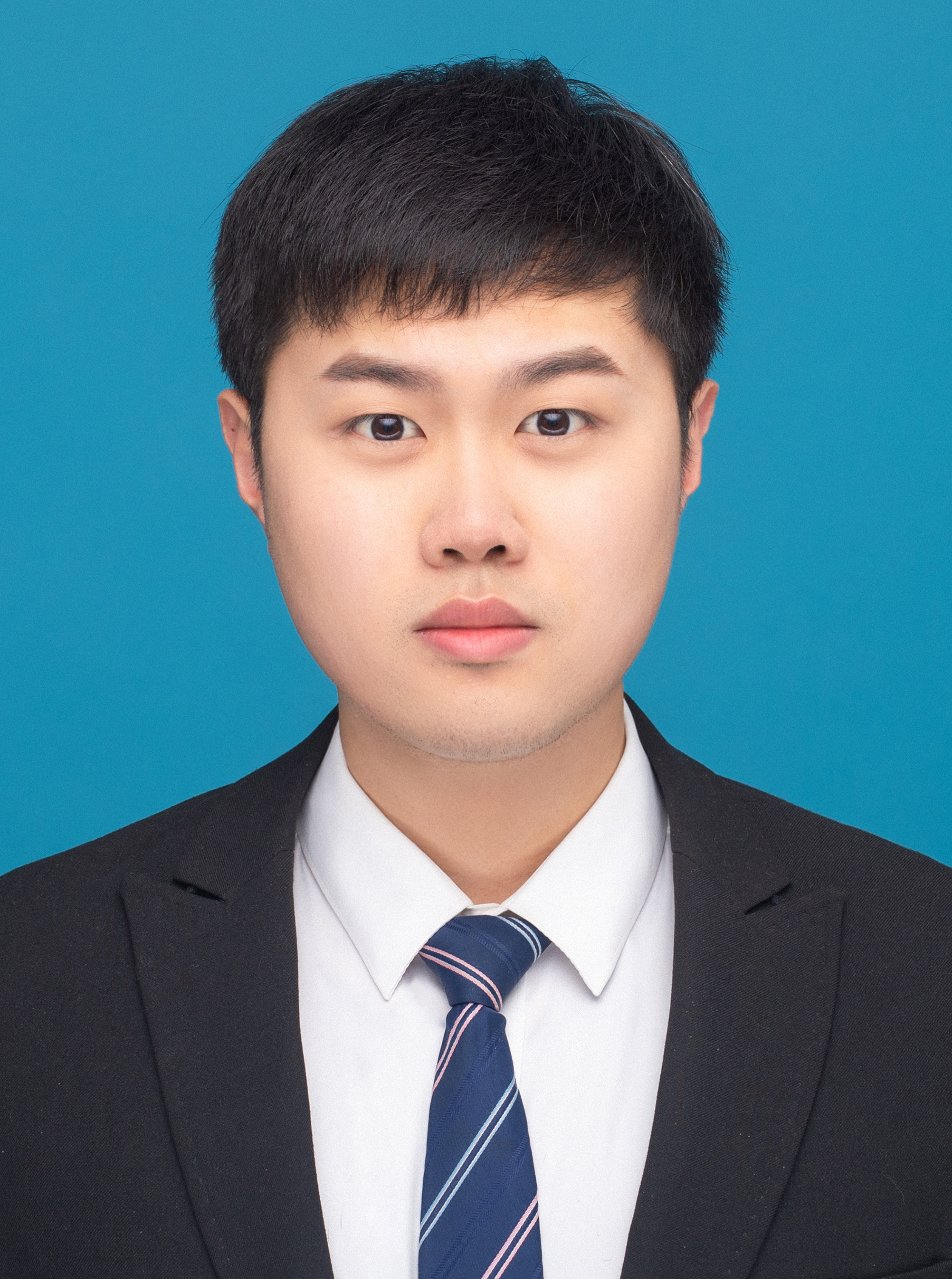}}]
	{Yang Jiao} received the B.S. degree from Central South University, Changsha, China, in 2020. He is currently working toward the Ph.D. degree in computer science with the Department of Computer Science and Technology, Tongji University, Shanghai, China. His research interests include machine learning, distributed optimization, and bilevel optimization.
\end{IEEEbiography}

\vfill

\end{document}